%% file: main.tex
\documentclass[runningheads]{llncs}
\usepackage[T1]{fontenc}
\usepackage{graphicx}
\usepackage{float}

\usepackage{paralist}
\usepackage{amsmath}
\usepackage{bm}
\usepackage{tikz}
\usepackage{booktabs}
\usepackage{multicol}
\usepackage{multirow}
\usepackage{orcidlink}
\usepackage{amssymb}

\usetikzlibrary{patterns,shapes,arrows,shapes.multipart}

\usepackage{xspace}
\newcommand{\R}{\textsuperscript{\textregistered}\xspace}

\begin{document}
\title{Benchmarking Deep Learning Convolutions on Energy-constrained CPUs}
%
%

\author{Enrique Galvez\inst{\orcidlink{0009-0004-9675-4049}} \and
Adrien Cassagne\inst{\orcidlink{0000-0002-6741-5329}} \and
Alix Munier\inst{\orcidlink{0000-0002-2170-6366}} \and
Manuel Bouyer}
\institute{Sorbonne Université, CNRS, LIP6, F-75005, Paris, France
}

%
%
%
\institute{}

\maketitle              
%
\begin{abstract}

This work evaluates State-of-the-Art convolution algorithms for CPU-based CNN inference. Although most prior studies focus on GPUs or NPUs, CPU implementations remain comparatively under-optimized. Our first contribution is to provide fair benchmarking for embedded CPU inference. 
We evaluate direct, GEMM-based, and Winograd convolutions across modern CPUs from ARM\R, Intel\R, AMD\R, and NVIDIA\R vendors, considering both latency and energy efficiency.
To the best of our knowledge, this is the first study to present a fair, cross-vendor comparison of CPU energy consumption using a high-resolution socket-level measurement platform. To validate our methodology, we further compare socket-level power measurements with estimates derived from model-specific registers (MSRs), finding that MSRs underestimate the power consumption of convolution inference by 10–30\%.
Our results show that the ARM\R Cortex-A78AE CPU combined with an implicit GEMM convolution implementation offers the best trade-off between latency and power consumption, achieving ResNet50v1.5 inference in 102~ms with an average power of 25.3~W, corresponding to 2.58~J.



\end{abstract}

\section{Introduction and Related work}

Deep neural networks (DNNs) have become pervasive in modern embedded computer vision systems. Among them, convolutional neural networks (CNNs) remain the backbone of most classification~\cite{Krizhevsky2017} and detection pipelines~\cite{Du_2018}, thanks to their relatively low computational and memory requirements as well as the maturity of available software frameworks~\cite{khanam2024yolov11overviewkeyarchitectural,10925383}. Even with the rise of Transformer-based models such as DETR~\cite{10.1007/978-3-030-58452-8_13}, CNN backbones still account for a significant portion of inference time: up to 30\% in some configurations~\cite{ma2023revisitingdetrpretrainingobject}. As a consequence, convolutions remain a central performance bottleneck in embedded inference of deep learning models.

The convolution operator is the key enabler of automatic feature extraction, but it also dominates the computational workload and energy demand during both training and inference~\cite{Sze2017}. In this paper, we show that convolutions account for more than 90\% of the inference of popular networks such as ResNet50v1.5, VGG19 and GoogLeNet (see Section~\ref{sec:protocol}). This cost has motivated extensive research into optimized implementations for GPUs~\cite{Lym2019,10.1007/978-3-030-61864-3_35,10.1145/3332466.3374520} and CPUs~\cite{
Dolz2022,Georganas2018,Santana2023}. However, most of these studies target high-performance hardware or specific architectures, and little attention has been given to embedded CPUs, which are still widely deployed in systems where accelerators are absent or offloading is impractical due to small models or input sizes.

In this article, we address this gap by systematically benchmarking the inference of CNNs on CPUs. We consider CPUs commonly used in embedded and battery-powered systems from the main vendors: ARM\R, Intel\R, AMD\R and NVIDIA\R . Our study focuses on three complementary performance metrics: latency, power and energy consumption. These metrics are particularly critical in embedded contexts where energy efficiency and thermal constraints directly affect system design. This study extends the work of Dolz et al.~\cite{Dolz2023}, which was limited to ARM-based CPUs. Furthermore, we introduce a protocol leveraging a novel high-resolution power measurement system~\cite{dalek}, enabling fine-grained characterization of the full system’s energy profile during convolution execution. This work includes the following key contributions:
\begin{inparaenum}[(i)]
  \item The characterization of a novel protocol based on high-frequency socket power measurement for benchmarking convolutions and the comparison of socket-measured power with power estimated using model-specific registers (MSRs); 
  \item A fair and cross-vendor evaluation of State-of-the-Art convolution algorithms on modern embedded CPUs, supported by a multidimensional benchmark considering both inference latency and power consumption in order to guide algorithmic and hardware choices.
\end{inparaenum}

The remainder of this article is organized as follows. Section~\ref{sec:algo} presents the convolution algorithms evaluated in this study. Section~\ref{sec:platform} describes the hardware and software platforms considered. Section~\ref{sec:protocol} outlines the experimental protocol. Section~\ref{sec:msrresults} compares socket-based power measurements with estimates obtained from MSRs. Section~\ref{sec:algresults} analyzes the energy consumption of the convolution algorithms across the studied CPUs. Section~\ref{sec:cnnresults} provides a benchmark of CNN inference across these CPUs. Finally, Section~\ref{sec:conclusion} concludes the article.



%
\section{Convolution algorithms}\label{sec:algo}


The convolution operation consists of applying a filter with learnable values across an image. We consider 2-dimensional convolutions over $4$-dimensional tensors, which are used by most CNNs in the context of computer vision. Moreover, since convolutions in such networks are rarely dilated or strided, we limit our study to undilated and unstrided convolutions. 

\vspace{-8mm}
\begin{table}[H]
   \caption{Notation for the main convolution parameters.}
   \centering
   \resizebox{.9\textwidth}{!}{
       \begin{tabular}{|l|l||l|l|}
       \hline
       MB     & batch size             & IH, IW & input height, width   \\
       KH, KW\phantom{a} & kernel height, width   & OH, OW \phantom{a} & output height, width  \\
       IC, OC & input, output channels \phantom{a} & PH, PW & padding height, width \phantom{a}  \\
       \hline
       \end{tabular}
   }
   \label{tab:notations}
\end{table}
\vspace{-5mm}

Using the notation described in Table \ref{tab:notations}, the dimensions of the input tensors can be written as: $MB \times IC \times IH \times IW$ and the dimensions of the weight tensors can be written as: $OC\times IC \times KH \times KW$. Following the same notations, the convolution operation can be described formally as:

\vspace{-4mm}

\begin{equation}
\label{eq:conv}
\begin{aligned}
dst&[mb, oc, oh, ow]
  = bias[oc] \\
  & + \displaystyle\sum_{ic=0}^{IC-1} \sum_{kh=0}^{KH-1} \sum_{kw=0}^{KW-1}
     src(mb, ic, ih, iw)\cdot weights[oc, ic, kh, kw],
\end{aligned}
\end{equation}

where $ih := oh + kh - PH$ and $iw := ow + kw - PW$. 

\vspace{3mm}

The output tensor $dst$ is computed given an input tensor $src$ and using learned values for $bias$ and $weights$ tensors. Each element of the output tensor is computed as a weighted sum of elements from the input tensor across the three dimensions of the kernel: channels ($IC$), kernel height ($KH$) and kernel width ($KW$). Figure \ref{fig:convolution} describes the convolution operator.
\begin{figure}[H]
    \vspace{-3mm}
    \centering
    \resizebox{.9\textwidth}{!}{ \input{sketch/direct.tex} }
    \caption{Convolution Product.}
    \label{fig:convolution}
\end{figure}


\vspace{-5mm}

In this article, we aim to minimize the latency of one CNN inference so we consider the batch size to be equal to 1. 
Predominant CNNs have convolution layers with $3\times 3$ kernels ($KH=KW=3$), $1\times 1$ kernels and, less commonly, bigger kernels such that $7\times 7$, for example. Later, we detail in Section~\ref{sec:protocol} the temporal impact of different convolutions on the inference of common CNNs. The $1\times 1$ convolutions can be implemented directly as a General Matrix Multiplication (GEMM), which is often preferred for performance. Other convolution kernels require specific implementations, discussed in the following paragraphs.
Our study proposes a benchmark of State-of-the-Art's three most commonly used implementations of convolutions in modern deployment of CNN inference: direct, GEMM-based and winograd, described as follows.  

\vspace{3mm}

\textbf{Direct method} [\texttt{direct}].
The convolution operator can be naively implemented using nested loops iterating through the output pixels and corresponding kernels. 
However, by reordering the loops and grouping the indices iterating over $[0,OC]$, $[0,IC]$, and $[0,OW]$ into fixed-size blocks, Zhang et al. \cite{Zhang2018} developed an optimized implementation of this computation, which we refer to as \texttt{direct}.

\vspace{3mm}

\textbf{Explicit and Implicit Lowering} [\texttt{im2row} and \texttt{gemm}].
These methods convert a convolution operation into a General Matrix Multiplication (GEMM) using ``lowering'' techniques. The principle is to replicate data in order to transform input and weight tensors into matrices whose product corresponds to the convolution output. 
This approach can be particularly efficient, as GEMM is well optimized on modern CPUs \cite{Kragstrom1998} and GPU-based accelerators \cite{Abdelfattah2016}. 
The lowering procedure is presented in Figure~\ref{fig:im2rowtransf} and we refer to our explicit lowering convolution as \texttt{im2row} in the following.
Although \texttt{im2row} involves a large overhead due to data movements, implicit lowering reduces this overhead by transforming small tiles of the tensor on-the-fly and computing the output result for each tile in parallel~\cite{Lym2019}. 
The default implementation of OneDNN~\cite{oneDNN} uses implicit lowering and is referred to as \texttt{gemm} in the following.
These methods are particularly efficient on highly parallel architectures such as GPUs or TPUs~\cite{Zhou2021}.

\vspace{-5mm}
\begin{figure}
    \centering
   \resizebox{\textwidth}{!}{ \input{sketch/im2row.tex} }
    \caption{Computing a convolution using \texttt{im2row}.}
    \label{fig:im2rowtransf}
\end{figure}
\vspace{-5mm}


\textbf{Winograd convolution} [\texttt{wino}].
The last method is an extension of Winograd's algorithm~\cite{Winograd1980} which reduces the number of floating-point multiplications of one-dimensional convolutions by taking advantage of redundancy patterns, at the cost of increasing the number of additions. To transpose this method to deep-learning convolutions, we followed the approach of Lavin and Gray~\cite{Lavin2015}, working on $3\times 3$ convolution kernels. 
This algorithm is referred to as \texttt{wino} and its main benefit is to reduce the number of critical floating-point multiplications by $2.25$ when the window size is 2.
Such methods were implemented on GPUs~\cite{Park2016}, but are known to show floating-point imprecision with quantized data \cite{Meng2019}.



\section{Experiments Platform}\label{sec:platform}

Our study targets CPUs typically deployed in embedded, battery-powered systems from the major processors vendors (ARM\R, Intel\R, AMD\R, NVIDIA\R). We selected the processors described in Table \ref{tab:archis} for executing our workloads.
The first class of CPUs considered in this study comprises the processors integrated into NVIDIA Jetson single-board computers, which are designed for computer vision. The CPUs of the AGX Orin and AGX Xavier boards both implement the ARM v8.2 instruction set: AGX Orin has a ARM Cortex-A78AE CPU and AGX Xavier has a NVIDIA Carmel CPU. A second class includes x86 processors. We evaluate a Ryzen 7 7840U based on the Zen 4 architecture and a Ryzen AI 9 HX 370 based on Zen 5. Finally, we benchmark an Intel Core Ultra 9 185H CPU from the Meteor Lake generation. Some of the studied architectures have heterogeneous CPUs, with performance cores, efficiency cores or low power efficiency cores (respectively p, e and LPe cores). p-cores are the fastest CPU cores but they are more energy consuming than e and LPe cores that are slower. Some CPUs also allow 2-way simultaneous multithreading (2-SMT) on some of their cores. All platforms use Ubuntu distribution. The corresponding Linux kernels are reported in Table~\ref{tab:archis}.

\setlength{\tabcolsep}{7pt}
\vspace{-8mm}
\begin{table}
    \caption{Description of the targets with $\mathcal{T}$ the core type and $\mathcal{C}$ the cores number.}
    \label{tab:archis}
    \vspace{0.1cm}
    {\resizebox{1.0\linewidth}{!}{{\Large
\begin{tabular}{c l c r r l r r r r r}
  \toprule
  & & & & \multicolumn{5}{c}{\text{CPU core}} & \multicolumn{2}{l}{Software version} \\ \cmidrule(lr){5-9} \cmidrule(lr){10-11}
                        &        & TDP &     &      &              &     &     & Proc. & Linux &     \\
  Product \& \emph{Tag} & Vendor & (Watts) & Release & $\mathcal{T}$ & Architecture & $\mathcal{C}$ & SMT &  (nm)  & (Ubuntu) & GCC \\
  \midrule
  \multirow{2}{*}{\shortstack[c]{Jetson AGX Xavier\\ RAM 16 GB \textit{(NV Carmel)}}} & \multirow{2}{*}{NVIDIA} & \multirow{2}{*}{30 $\leq$} & \multirow{2}{*}{Oct'18} &   \multirow{2}{*}{p} & \multirow{2}{*}{Carmel}       &  \multirow{2}{*}{8} & \multirow{2}{*}{1} & \multirow{2}{*}{12} & \multirow{2}{*}{4.09.201} & \multirow{2}{*}{11.4.0} \\ 
                                                                            &                         &                            &                         &                      &                               &                     &                    &                     &                           &                         \\ \addlinespace
  \multirow{2}{*}{\shortstack[c]{Jetson AGX Orin\\ RAM 64 GB \textit{(ARM Cortex)}}}  & \multirow{2}{*}{ARM}    & \multirow{2}{*}{60 $\leq$} & \multirow{2}{*}{Mar'23} &   \multirow{2}{*}{p} & \multirow{2}{*}{Cortex-A78AE} & \multirow{2}{*}{12} & \multirow{2}{*}{1} &  \multirow{2}{*}{8} & \multirow{2}{*}{5.10.120} & \multirow{2}{*}{11.4.0} \\
                                                                            &                         &                            &                         &                      &                               &                     &                    &                     &                           &                         \\ \addlinespace
  \multirow{2}{*}{\shortstack[c]{Ryzen 7 7840U\\ RAM 32 GB \textit{(AMD Zen4)}}}      & \multirow{2}{*}{AMD}    & \multirow{2}{*}{15--30}    & \multirow{2}{*}{Jan'24} &   \multirow{2}{*}{p} & \multirow{2}{*}{Zen 4}        &  \multirow{2}{*}{8} & \multirow{2}{*}{2} &  \multirow{2}{*}{5} & \multirow{2}{*}{6.08.000} & \multirow{2}{*}{13.3.0} \\
                                                                            &                         &                            &                         &                      &                               &                     &                    &                     &                           &                         \\ \addlinespace
  \multirow{3}{*}{\shortstack[c]{Core Ultra 9 185H\\ RAM 32 GB \textit{(Intel U9)}}}  & \multirow{3}{*}{Intel}  & \multirow{2}{*}{35--115}   & \multirow{3}{*}{Jul'24} &                 LPe  &                 Crestmont     &                  2  &                 1  &                  5  & \multirow{3}{*}{6.14.000} & \multirow{3}{*}{14.2.0} \\ 
                                                                            &                         &                            &                         &                   e  &                 Crestmont     &                  8  &                 1  &                  7  &                           &                         \\ 
                                                                            &                         &                            &                         &                   p  &                 Redwood Cove  &                  6  &                 2  &                  7  &                           &                         \\ \addlinespace 
  \multirow{2}{*}{\shortstack[c]{Ryzen AI 9 HX 370 32 GB\\ \textit{(AMD Zen5)}}}  & \multirow{2}{*}{AMD}    & \multirow{2}{*}{15--54}    & \multirow{2}{*}{Oct'24} &                   e  &                 Zen 5c        &                  8  &                 2  &                  4  & \multirow{2}{*}{6.08.000} & \multirow{2}{*}{13.3.0} \\
                                                                            &                         &                            &                         &                   p  &                 Zen 5         &                  4  &                 2  &                  4  &                           &                         \\
  \bottomrule
\end{tabular}
    }}}
    \vspace{-5mm}
\end{table}

The algorithms described in Section \ref{sec:algo} are implemented into Intel's OneDNN framework version 3.4 \cite{oneDNN}. This framework provides an abstraction level that allows one to implement deep learning primitives, with a built-in correctness and performance evaluation tool called BenchDNN. We compiled OneDNN with the \texttt{-fopenmp -O3} optimization flags. OpenMP runs threads to leverage tensor parallelism. 
The measurements have been first conducted using BenchDNN, focusing only on the convolutions. Later, a full inference across the ResNet50v1.5 network~\cite{He2015} is evaluated. Its execution relies on ONNX Runtime that links with our OneDNN implementations. ONNX Runtime v1.22.2 was used for all architectures except for NV Carmel which used version 1.17.0 for compatibility. It is worth mentioning that inter-operation parallelism is disabled. Only tensor parallelism inside OneDNN convolutions remains.



\section{Experiments Protocol}\label{sec:protocol}

\begin{figure}[H]
    \centering
    \includegraphics[width=\textwidth]{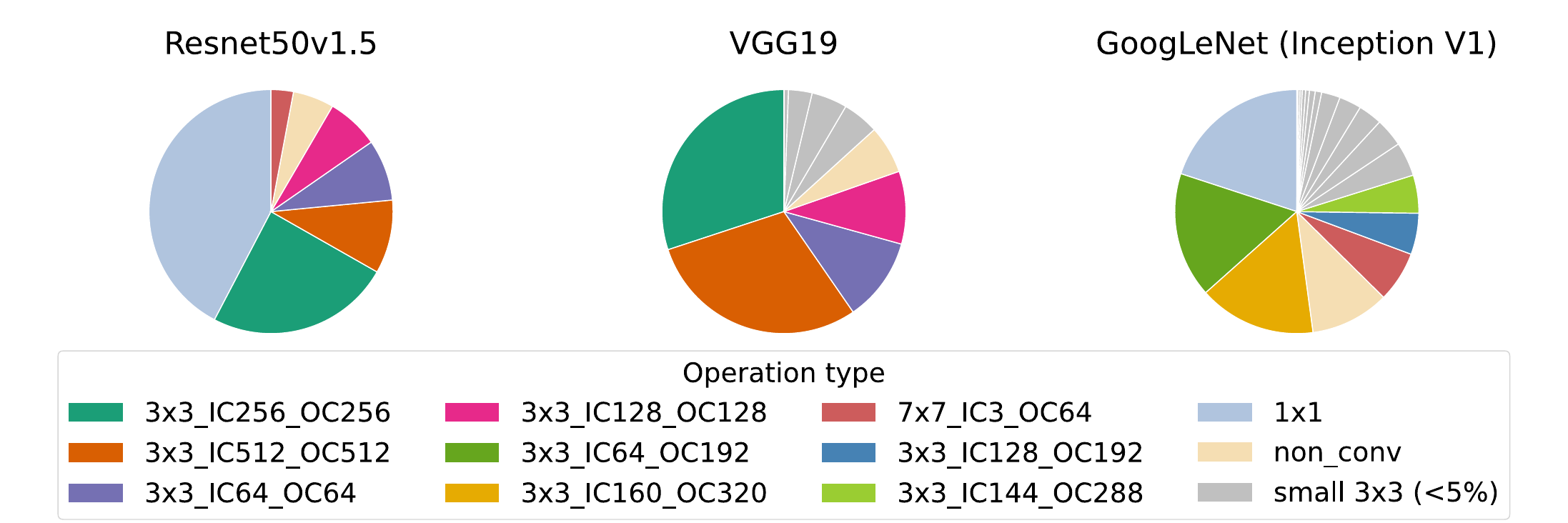}    
    \caption{Impact of relative latency of convolutions on CNN inference depending on their dimensions.}
    \label{fig:comp_algscamembert}
\end{figure}

The aim of this study is to benchmark latency and energy consumption of CPUs from several vendors for the inference of CNNs. An important thing to note in that we consider both convolution-level benchmark using BenchDNN and network-level inference benchmark using ONNX Runtime. 
In this context, Figure \ref{fig:comp_algscamembert} compares the temporal cost of several convolutions over the inference of three popular CNN networks: ResNet50v1.5, VGG19 and GoogLeNet. The results have been obtained using ONNX Runtime built-in profiler on the AMD Ryzen7 7840U CPU. In order to measure per-layer timings, we set ONNX Runtime's \texttt{GraphOptimizationLevel} to \texttt{ORT\_DISABLE\_ALL} to disable layer fusion.

A first observation we can make from Figure \ref{fig:comp_algscamembert} is that convolutions clearly dominate inference time of evaluated CNNs: often more than 90\% of inference time. Another observation is that $1\times 1$ convolutions cost less than $3\times 3$ convolutions individually but the high number of $1\times 1$ convolutions in ResNet makes their cost non-negligible (42\% of inference time). Despite the high cost of $1\times 1$ convolutions in ResNet50v1.5, $3\times 3$ convolutions cost approximately 52\% of inference time. Another interesting observation is that predominant $3\times 3$ convolutions have similar kernels between the networks. In particular, the most expensive convolutions of ResNet50v1.5 and VGG19 share the same dimensions. As a consequence to these observations, our convolution-level benchmarking will only focus on the most expensive $3\times 3$ convolution in terms of computations: \texttt{MB1\_IC256IH14\_OC256OH14\_KH3PH1} (which we refer to as \texttt{3x3\_IC256\_OC256} in Figure \ref{fig:comp_algscamembert} and in the following). In the previous description format, each convolution parameter (from Table~\ref{tab:notations}) is followed by its value. For instance, \texttt{MB1} means that minibatch is 1 corresponding to inference without batching.

\vspace{-7mm}

\vspace{-7mm}

One of the contributions of this work is the experimental methodology. 
While its steps are not new on their own, the novelty of our approach resides in the use, for the first time, of high frequency measurement boards for monitoring socket power consumption of convolutions. 
The key methodology steps are:
\vspace{-1mm}
\begin{enumerate}
    \item Using \texttt{taskset} system command, we run the main program on a specific set of CPU HW threads (compact threads pinning on cores or HW threads).
    \item We run a warm-up round of 200 convolutions or inferences, in order to ensure the stability of execution speed (stabilization of the CPU frequency).
    \item We create a background thread that collects the energy measurements. It is pinning on a unused core when possible.
    \item We run 1000 iterations of convolutions or inferences and gather the speed metrics returned by BenchDNN or ONNX Runtime
    \item We stop the background thread and compute statistics over the energy measurements. 
\end{enumerate}

We used this methodology for all the following measurements, to ensure their pertinence and reproducibility. Section \ref{sec:msrresults} analyzes the pertinence of socket measurements and provide important insights over the measurements uncertainty that can be expected from experimental results.

\section{MSRs versus Socket Measure}\label{sec:msrresults}



To motivate our choice of high-frequency socket measurements over MSRs for power evaluation, we begin by comparing both methods. A first difference resides in the way that power is estimated in both cases. In one hand, MSR power is estimated by reading a register containing the energy consumption of the exposed power domain. 
By reading these registers at a certain frequency, 
the average power consumption is estimated for each interval. On the other hand, our high-frequency measurement system measures energy consumption between the power supply and the socket at 1000~samples/s for x86 CPUs~\cite{dalek} or 5000~samples/s for Jetson boards.

\vspace{2mm}

In order to evaluate the differences between MSRs power estimates and socket power measurements in the context of convolutions, we performed rounds of experiments for the \texttt{3x3\_IC256\_OC256} convolution, following the protocol described in Section \ref{sec:protocol}. Since NVIDIA provides \texttt{jetson-power-tools} to precisely report power consumption across the components of Jetson boards, we used it to gather MSR values at 10 samples/s. For x86 CPUs, MSR power is gathered using \texttt{perf stat} to read the RAPL \texttt{/power/energy-pkg} event at 1 sample/s. We chose these sampling rates to minimize errors due to measurement overhead and the lack of energy updates \cite{raplinaction}.

\vspace{-7mm}

\begin{figure}[H]
   \centering
   \includegraphics[width=\textwidth]{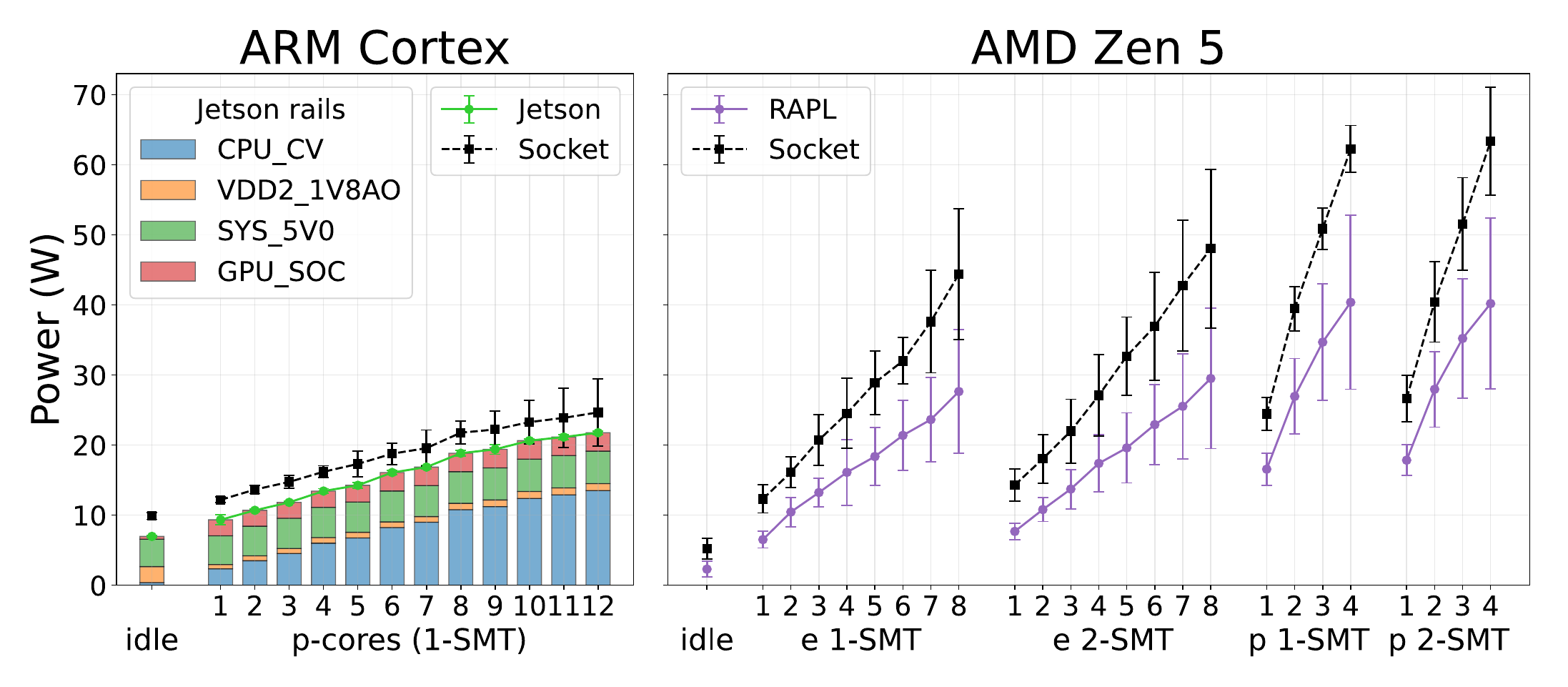}    
   \caption{MSRs versus socket measure of average power and dispersion (standard deviation) depending on multithreading.}
   \label{fig:powercompmt}
\end{figure}

\vspace{-5mm}

Figure \ref{fig:powercompmt} provides a comparison between MSRs-based and socket-based power measurements for the inference of the \texttt{3x3\_IC256\_OC256} convolution for ARM Cortex A78AE (Jetson Orin) and AMD Ryzen AI 9 HX370 (x86) CPUs.
A first observation we can make is that in both cases, measured MSR power is lower than socket power. However, the difference between socket power and Jetson power does not depend on multithreading while the difference between RAPL and socket power increases with multithreading.
A possible explanation for RAPL's behavior is the fact that RAM is not part of the exposed power domain. As RAM usage increases with parallelism, this may explain the increasing difference between RAPL and socket measurements for AMD Zen5. In addition to RAM, RAPL does not capture power supply losses, fans, storage or network, which may result in a non-constant overhead \cite{CompraplpowJay}. A last observation is that standard deviation of power measurements is high for both MSRs and socket readings. 




\vspace{-5mm}
 
\begin{figure}[H]
   \centering
   \includegraphics[width=\textwidth]{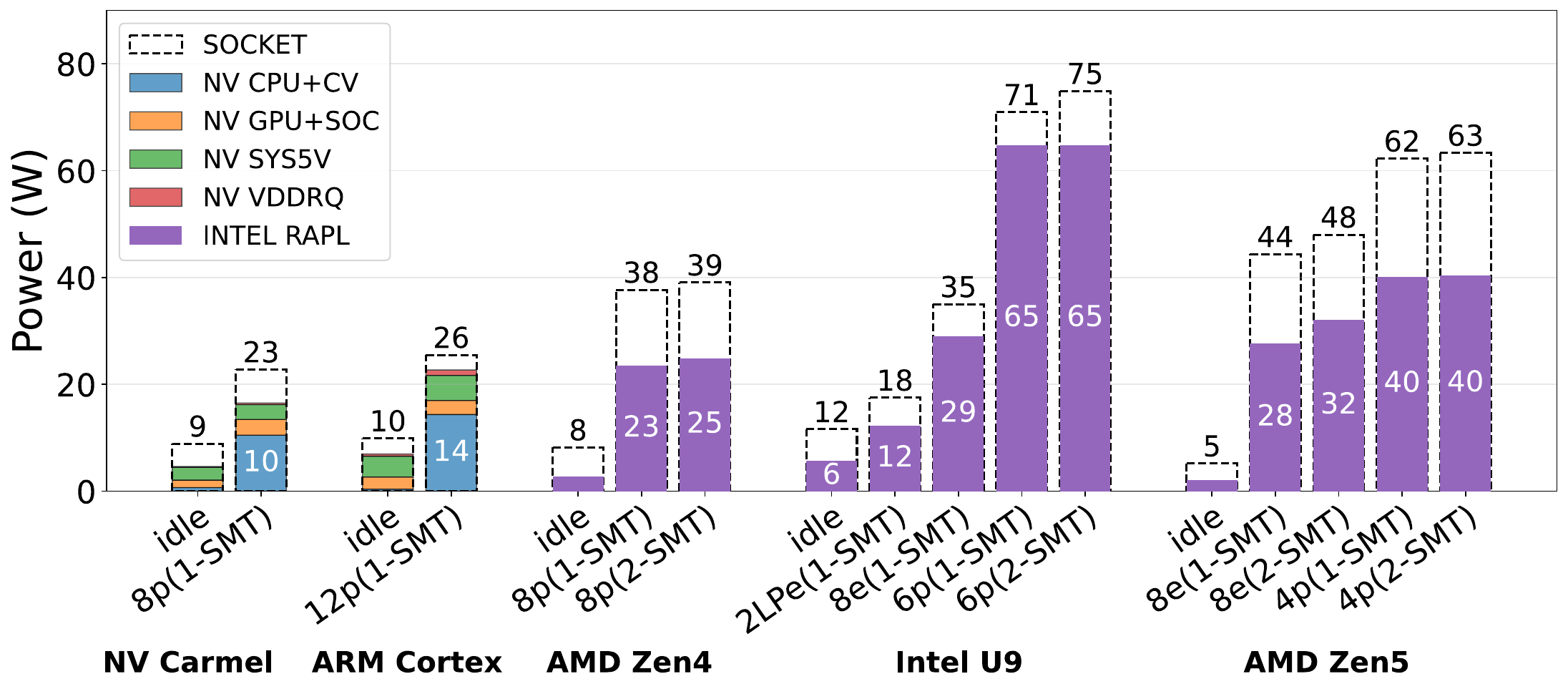}    
   \caption{Energy consumption measurements: MSRs and socket.}
   \label{fig:powercomp}
\end{figure}

\vspace{-5mm}


Using the same protocol, we extended the results displayed in Figure~\ref{fig:powercompmt} to all studied architectures in Figure~\ref{fig:powercomp} the difference between MSRs and socket power measurements. 
Based on these experiments, we conclude that MSRs-based power measurements do not fully capture the total power consumption measured at the socket. In idle state, MSRs readings are more than 50\% lower than the corresponding socket measurements, while during convolution computations, this discrepancy ranges between 10\% and 30\%.
Another observation is that 2-SMT have a small and positive impact over average power consumption.

One may notice that \texttt{jetson-power-tools} provides the closest approximation of the actual power consumption (compared to RAPL). Despite the high dispersion of power measurements for both MSRs and socket-based readings, the mean power is more consistent across repeated socket measurements than across repeated MSRs measurements. Consequently, for the following experiments of this article, we rely exclusively on socket measurements to accurately benchmark the power consumption of each system.




\section{Convolution performance depending on algorithm}\label{sec:algresults}

\begin{figure}[H]
    \centering
    \includegraphics[width=\textwidth]{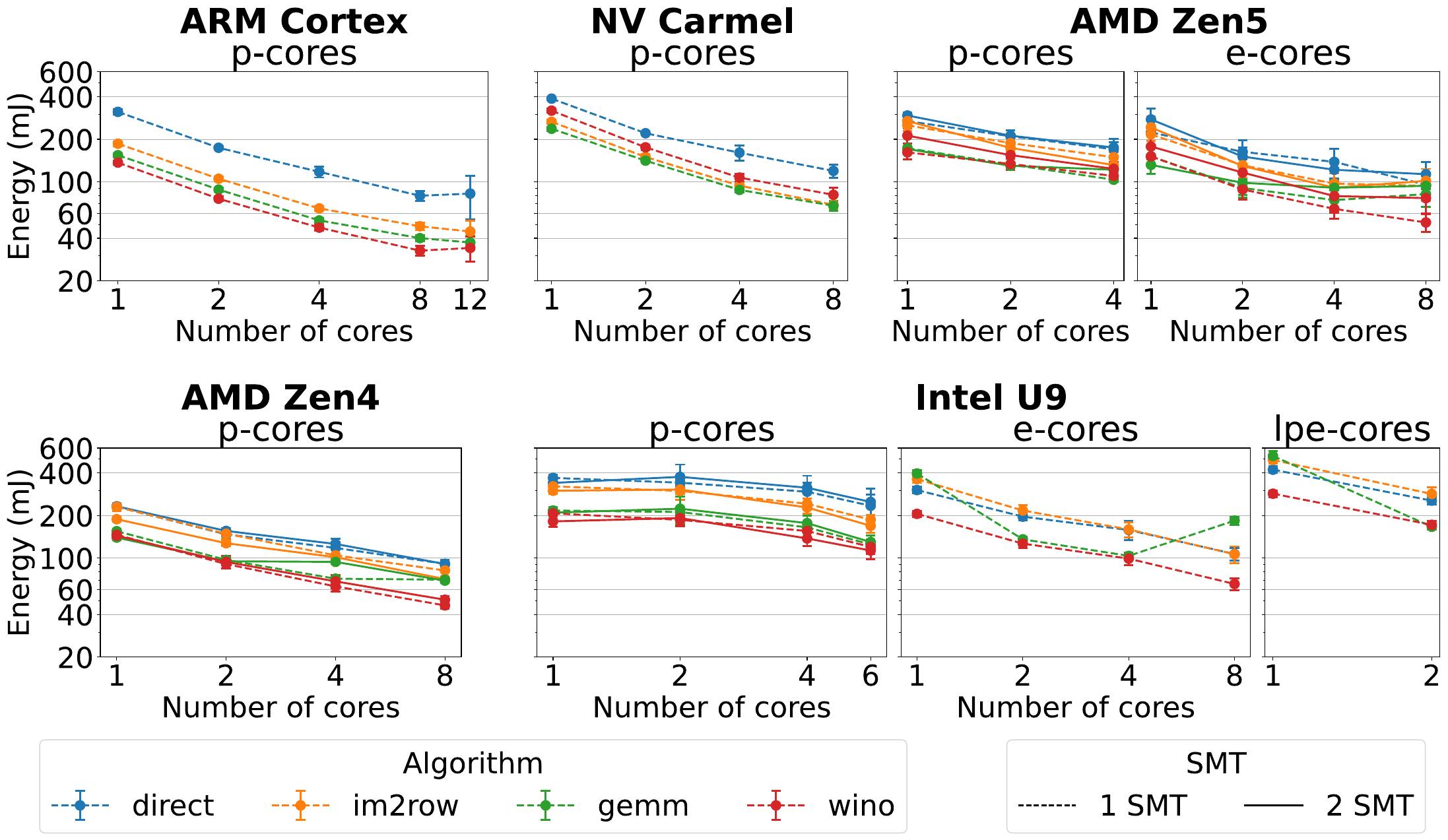}    
    \caption{Energy consumption of \texttt{3x3\_IC256\_OC256} convolution depending on algorithm and multithreading.}
    \label{fig:comp_algs}
\end{figure}

Figure \ref{fig:comp_algs} presents the energy consumed to compute the \texttt{3x3\_IC256\_OC256} convolution depending on the number of CPU cores. Additionally, Figure \ref{fig:comp_algs} also compares the different algorithms. An important remark is that uncertainty propagated to energy is way smaller than uncertainty over power, due to the low variance of latency measurements.

\vspace{-5mm}

\vspace{-7mm}





First, we observe that increasing the number of physical cores almost always decrease the energy consumption of the convolution. The reason is that increasing the number of cores further decreases latency than it increases power consumption. 
We also observe that 2-SMT has no significant impact on the average energy consumption of an operation. The reason is that convolution kernels already saturate the processor’s backend execution units. SMT, which mainly helps when backend resources are underutilized, does not provide additional throughput in this context.
A surprising result is that e-cores and LPe-cores consumes more than p-cores for the heterogeneous architectures. The reason is that low-power processors are much slower than they are energy efficient. 
Regarding the algorithms, \texttt{wino} and \texttt{gemm} are the most energy efficient because they are the fastest on almost every architecture.
The main result of this benchmark is that the best algorithm-architecture configurations use 8 p-cores of the ARM Cortex A78AE or the AMD Zen 4, with the \texttt{wino} algorithm. Both configurations reach a power consumption that is less than 60~mJ.

%

\section{Cross-vendor CPU benchmark for CNN inference}\label{sec:cnnresults}

\begin{figure}[H]
    \centering
    \includegraphics[width=\textwidth]{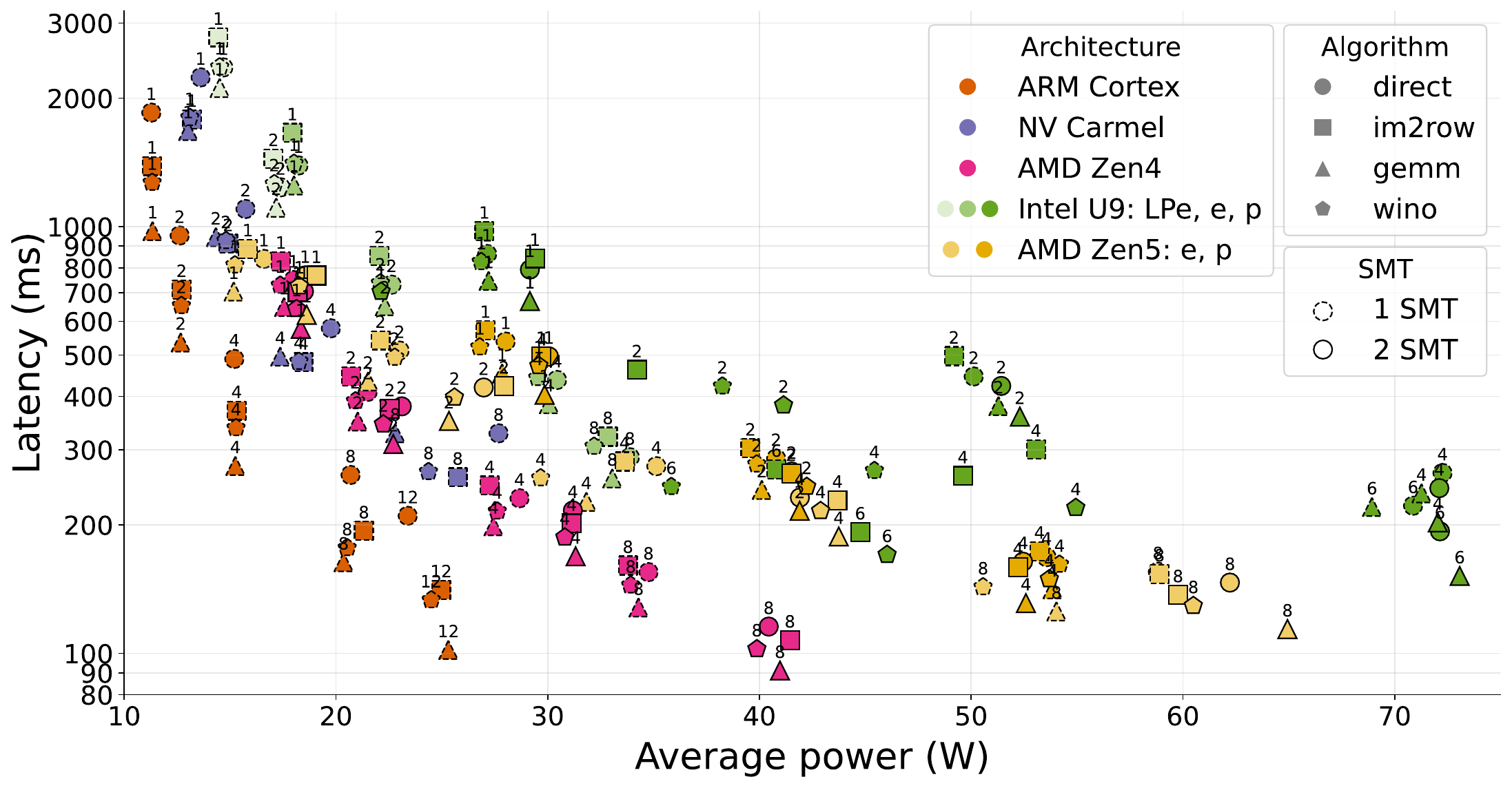}    
    \caption{Latency and power consumption depending on architecture, multithreading and algorithm for a full inference of ResNet50v1.5 .}
    \label{fig:pareto}
\end{figure}

Results shown in Figure \ref{fig:pareto} are measured through a full inference of ResNet50v1.5. Since we target energy constrained devices, power consumption may be a discriminant criteria for the architectures, as well as inference latency. All the inference runs are executed on the same $640\times640$ image from COCO dataset~\cite{cocodataset} and we ensured the correctness of the predictions. 

\vspace{-3mm}

\vspace{-5mm}

A first observation is that focusing on the computational part of the convolution allows \texttt{wino} to take advantage of its arithmetic complexity optimization (see Figure \ref{fig:comp_algs}) while in a full inference, significant data management acts in favour of implicit GEMM (\texttt{gemm}) implementation. In addition to that, Figure \ref{fig:pareto} clearly shows the existence of a trade-off between latency and average power consumption in our context. Moreover, among the studied architecture, ARM Cortex A78AE is the architecture offering the best trade-off between inference latency and average power consumption. If power budget is higher, AMD Zen4 can allow a faster inference. Additionally, NVIDIA Carmel CPU allow slow inference with a low power budget while, on the contrary, AMD Zen5 offers good latency at the cost of high power consumption. However, AtomMan X7 Ti offers sub-optimal performances on each one of its CPU coretypes.

\section{Conclusion}\label{sec:conclusion}


This work introduces an original methodology based on high-resolution socket-level power measurements to benchmark CNN inference on CPUs commonly used in embedded and battery-powered systems. By extending prior studies to the major CPU vendors ARM\R, Intel\R, AMD\R, NVIDIA\R and leveraging precise socket-level energy instrumentation, we provide accurate insights into the real energy consumption of such CPUs. Our results demonstrate the relevance of socket-level energy measurements, which we prove to be more reliable and precise than MSR-based estimations. Our experiments also reveal a clear trade-off between inference latency and power consumption across all evaluated architectures. Overall, our benchmarks show that the ARM\R Cortex-A78AE CPU combined with an implicit GEMM convolution implementation offers the best trade-off between latency and power consumption, achieving ResNet50v1.5 inference in 102~ms with an average power of 25.3~W, corresponding to 2.58~J. Future work will expand this analysis to architectures with accelerators such as GPUs or NPUs, and further investigate the discrepancy between MSRs and socket energy measurements.


\bibliographystyle{splncs04}
\bibliography{biblio}

\end{document}

%% file: sketch/direct.tex
 
\tikzset{
pattern size/.store in=\mcSize, 
pattern size = 5pt,
pattern thickness/.store in=\mcThickness, 
pattern thickness = 0.3pt,
pattern radius/.store in=\mcRadius, 
pattern radius = 1pt}
\makeatletter
\pgfutil@ifundefined{pgf@pattern@name@_rbrybacpb}{
\pgfdeclarepatternformonly[\mcThickness,\mcSize]{_rbrybacpb}
{\pgfqpoint{0pt}{-\mcThickness}}
{\pgfpoint{\mcSize}{\mcSize}}
{\pgfpoint{\mcSize}{\mcSize}}
{
\pgfsetcolor{\tikz@pattern@color}
\pgfsetlinewidth{\mcThickness}
\pgfpathmoveto{\pgfqpoint{0pt}{\mcSize}}
\pgfpathlineto{\pgfpoint{\mcSize+\mcThickness}{-\mcThickness}}
\pgfusepath{stroke}
}}
\makeatother

 
\tikzset{
pattern size/.store in=\mcSize, 
pattern size = 5pt,
pattern thickness/.store in=\mcThickness, 
pattern thickness = 0.3pt,
pattern radius/.store in=\mcRadius, 
pattern radius = 1pt}
\makeatletter
\pgfutil@ifundefined{pgf@pattern@name@_s09s8eclh}{
\pgfdeclarepatternformonly[\mcThickness,\mcSize]{_s09s8eclh}
{\pgfqpoint{0pt}{-\mcThickness}}
{\pgfpoint{\mcSize}{\mcSize}}
{\pgfpoint{\mcSize}{\mcSize}}
{
\pgfsetcolor{\tikz@pattern@color}
\pgfsetlinewidth{\mcThickness}
\pgfpathmoveto{\pgfqpoint{0pt}{\mcSize}}
\pgfpathlineto{\pgfpoint{\mcSize+\mcThickness}{-\mcThickness}}
\pgfusepath{stroke}
}}
\makeatother

 
\tikzset{
pattern size/.store in=\mcSize, 
pattern size = 5pt,
pattern thickness/.store in=\mcThickness, 
pattern thickness = 0.3pt,
pattern radius/.store in=\mcRadius, 
pattern radius = 1pt}
\makeatletter
\pgfutil@ifundefined{pgf@pattern@name@_ip5irdekh}{
\pgfdeclarepatternformonly[\mcThickness,\mcSize]{_ip5irdekh}
{\pgfqpoint{0pt}{0pt}}
{\pgfpoint{\mcSize+\mcThickness}{\mcSize+\mcThickness}}
{\pgfpoint{\mcSize}{\mcSize}}
{
\pgfsetcolor{\tikz@pattern@color}
\pgfsetlinewidth{\mcThickness}
\pgfpathmoveto{\pgfqpoint{0pt}{0pt}}
\pgfpathlineto{\pgfpoint{\mcSize+\mcThickness}{\mcSize+\mcThickness}}
\pgfusepath{stroke}
}}
\makeatother

 
\tikzset{
pattern size/.store in=\mcSize, 
pattern size = 5pt,
pattern thickness/.store in=\mcThickness, 
pattern thickness = 0.3pt,
pattern radius/.store in=\mcRadius, 
pattern radius = 1pt}
\makeatletter
\pgfutil@ifundefined{pgf@pattern@name@_oy72wmuew}{
\pgfdeclarepatternformonly[\mcThickness,\mcSize]{_oy72wmuew}
{\pgfqpoint{0pt}{0pt}}
{\pgfpoint{\mcSize+\mcThickness}{\mcSize+\mcThickness}}
{\pgfpoint{\mcSize}{\mcSize}}
{
\pgfsetcolor{\tikz@pattern@color}
\pgfsetlinewidth{\mcThickness}
\pgfpathmoveto{\pgfqpoint{0pt}{0pt}}
\pgfpathlineto{\pgfpoint{\mcSize+\mcThickness}{\mcSize+\mcThickness}}
\pgfusepath{stroke}
}}
\makeatother

 
\tikzset{
pattern size/.store in=\mcSize, 
pattern size = 5pt,
pattern thickness/.store in=\mcThickness, 
pattern thickness = 0.3pt,
pattern radius/.store in=\mcRadius, 
pattern radius = 1pt}
\makeatletter
\pgfutil@ifundefined{pgf@pattern@name@_7el980ygu}{
\pgfdeclarepatternformonly[\mcThickness,\mcSize]{_7el980ygu}
{\pgfqpoint{0pt}{-\mcThickness}}
{\pgfpoint{\mcSize}{\mcSize}}
{\pgfpoint{\mcSize}{\mcSize}}
{
\pgfsetcolor{\tikz@pattern@color}
\pgfsetlinewidth{\mcThickness}
\pgfpathmoveto{\pgfqpoint{0pt}{\mcSize}}
\pgfpathlineto{\pgfpoint{\mcSize+\mcThickness}{-\mcThickness}}
\pgfusepath{stroke}
}}
\makeatother

 
\tikzset{
pattern size/.store in=\mcSize, 
pattern size = 5pt,
pattern thickness/.store in=\mcThickness, 
pattern thickness = 0.3pt,
pattern radius/.store in=\mcRadius, 
pattern radius = 1pt}
\makeatletter
\pgfutil@ifundefined{pgf@pattern@name@_g2c110dgk}{
\pgfdeclarepatternformonly[\mcThickness,\mcSize]{_g2c110dgk}
{\pgfqpoint{0pt}{0pt}}
{\pgfpoint{\mcSize+\mcThickness}{\mcSize+\mcThickness}}
{\pgfpoint{\mcSize}{\mcSize}}
{
\pgfsetcolor{\tikz@pattern@color}
\pgfsetlinewidth{\mcThickness}
\pgfpathmoveto{\pgfqpoint{0pt}{0pt}}
\pgfpathlineto{\pgfpoint{\mcSize+\mcThickness}{\mcSize+\mcThickness}}
\pgfusepath{stroke}
}}
\makeatother
\tikzset{every picture/.style={line width=0.75pt}} 

\begin{tikzpicture}[x=0.75pt,y=0.75pt,yscale=-1,xscale=1]

\draw  [draw opacity=0] (870,130) -- (970,130) -- (970,230) -- (870,230) -- cycle ; \draw   (870,130) -- (870,230)(889.9,130) -- (889.9,230)(909.8,130) -- (909.8,230)(929.7,130) -- (929.7,230)(949.6,130) -- (949.6,230)(969.5,130) -- (969.5,230) ; \draw   (870,130) -- (970,130)(870,149.9) -- (970,149.9)(870,169.8) -- (970,169.8)(870,189.7) -- (970,189.7)(870,209.6) -- (970,209.6)(870,229.5) -- (970,229.5) ; \draw    ;
\draw  [draw opacity=0][fill={rgb, 255:red, 255; green, 255; blue, 255 }  ,fill opacity=1 ] (840,160) -- (939.5,160) -- (939.5,259.5) -- (840,259.5) -- cycle ;
\draw  [draw opacity=0][fill={rgb, 255:red, 255; green, 232; blue, 89 }  ,fill opacity=1 ] (144,60) -- (204,60) -- (204,120) -- (144,120) -- cycle ;
\draw  [draw opacity=0][dash pattern={on 4.5pt off 4.5pt}] (144,60) -- (343.9,60) -- (343.9,260) -- (144,260) -- cycle ; \draw  [color={rgb, 255:red, 155; green, 155; blue, 155 }  ,draw opacity=1 ][dash pattern={on 4.5pt off 4.5pt}] (144,60) -- (144,260)(163.9,60) -- (163.9,260)(183.8,60) -- (183.8,260)(203.7,60) -- (203.7,260)(223.6,60) -- (223.6,260)(243.5,60) -- (243.5,260)(263.4,60) -- (263.4,260)(283.3,60) -- (283.3,260)(303.2,60) -- (303.2,260)(323.1,60) -- (323.1,260)(343,60) -- (343,260) ; \draw  [color={rgb, 255:red, 155; green, 155; blue, 155 }  ,draw opacity=1 ][dash pattern={on 4.5pt off 4.5pt}] (144,60) -- (343.9,60)(144,79.9) -- (343.9,79.9)(144,99.8) -- (343.9,99.8)(144,119.7) -- (343.9,119.7)(144,139.6) -- (343.9,139.6)(144,159.5) -- (343.9,159.5)(144,179.4) -- (343.9,179.4)(144,199.3) -- (343.9,199.3)(144,219.2) -- (343.9,219.2)(144,239.1) -- (343.9,239.1)(144,259) -- (343.9,259) ; \draw  [color={rgb, 255:red, 155; green, 155; blue, 155 }  ,draw opacity=1 ][dash pattern={on 4.5pt off 4.5pt}]  ;
\draw  [draw opacity=0] (164,80) -- (324,80) -- (324,240) -- (164,240) -- cycle ; \draw   (164,80) -- (164,240)(183.9,80) -- (183.9,240)(203.8,80) -- (203.8,240)(223.7,80) -- (223.7,240)(243.6,80) -- (243.6,240)(263.5,80) -- (263.5,240)(283.4,80) -- (283.4,240)(303.3,80) -- (303.3,240)(323.2,80) -- (323.2,240) ; \draw   (164,80) -- (324,80)(164,99.9) -- (324,99.9)(164,119.8) -- (324,119.8)(164,139.7) -- (324,139.7)(164,159.6) -- (324,159.6)(164,179.5) -- (324,179.5)(164,199.4) -- (324,199.4)(164,219.3) -- (324,219.3)(164,239.2) -- (324,239.2) ; \draw    ;
\draw  [draw opacity=0][fill={rgb, 255:red, 255; green, 255; blue, 255 }  ,fill opacity=1 ] (94,128.7) -- (253.2,128.7) -- (253.2,288.2) -- (94,288.2) -- cycle ;
\draw  [draw opacity=0][fill={rgb, 255:red, 255; green, 255; blue, 255 }  ,fill opacity=1 ] (570,120) -- (629.7,120) -- (629.7,180) -- (570,180) -- cycle ;
\draw  [draw opacity=0] (570,120) -- (630,120) -- (630,180) -- (570,180) -- cycle ; \draw   (570,120) -- (570,180)(589.9,120) -- (589.9,180)(609.8,120) -- (609.8,180)(629.7,120) -- (629.7,180) ; \draw   (570,120) -- (630,120)(570,139.9) -- (630,139.9)(570,159.8) -- (630,159.8)(570,179.7) -- (630,179.7) ; \draw    ;
\draw  [draw opacity=0][pattern=_rbrybacpb,pattern size=6pt,pattern thickness=0.75pt,pattern radius=0pt, pattern color={rgb, 255:red, 189; green, 16; blue, 224}][dash pattern={on 6.75pt off 4.5pt}][line width=2.25]  (570.35,120.7) -- (629.7,120.7) -- (629.7,179.7) -- (570.35,179.7) -- cycle ;
\draw  [draw opacity=0][fill={rgb, 255:red, 255; green, 255; blue, 255 }  ,fill opacity=1 ] (540,150) -- (599.7,150) -- (599.7,210) -- (540,210) -- cycle ;
\draw  [draw opacity=0] (540,150) -- (600,150) -- (600,210) -- (540,210) -- cycle ; \draw   (540,150) -- (540,210)(559.9,150) -- (559.9,210)(579.8,150) -- (579.8,210)(599.7,150) -- (599.7,210) ; \draw   (540,150) -- (600,150)(540,169.9) -- (600,169.9)(540,189.8) -- (600,189.8)(540,209.7) -- (600,209.7) ; \draw    ;
\draw  [draw opacity=0][pattern=_s09s8eclh,pattern size=6pt,pattern thickness=0.75pt,pattern radius=0pt, pattern color={rgb, 255:red, 189; green, 16; blue, 224}][dash pattern={on 6.75pt off 4.5pt}][line width=2.25]  (540.35,150.7) -- (599.7,150.7) -- (599.7,209.7) -- (540.35,209.7) -- cycle ;
\draw  [draw opacity=0][fill={rgb, 255:red, 255; green, 255; blue, 255 }  ,fill opacity=1 ] (490.5,200) -- (550,200) -- (550,259.5) -- (490.5,259.5) -- cycle ;
\draw  [draw opacity=0] (490,200) -- (550,200) -- (550,260) -- (490,260) -- cycle ; \draw   (490,200) -- (490,260)(509.9,200) -- (509.9,260)(529.8,200) -- (529.8,260)(549.7,200) -- (549.7,260) ; \draw   (490,200) -- (550,200)(490,219.9) -- (550,219.9)(490,239.8) -- (550,239.8)(490,259.7) -- (550,259.7) ; \draw    ;
\draw  [draw opacity=0][pattern=_ip5irdekh,pattern size=6pt,pattern thickness=0.75pt,pattern radius=0pt, pattern color={rgb, 255:red, 65; green, 117; blue, 5}][dash pattern={on 6.75pt off 4.5pt}][line width=2.25]  (490,200) -- (550,200) -- (550,260) -- (490,260) -- cycle ;
\draw  [draw opacity=0][fill={rgb, 255:red, 255; green, 255; blue, 255 }  ,fill opacity=1 ] (460.5,230) -- (520,230) -- (520,289.5) -- (460.5,289.5) -- cycle ;
\draw  [fill={rgb, 255:red, 255; green, 232; blue, 89 }  ,fill opacity=1 ] (870,130) -- (890,130) -- (890,150) -- (870,150) -- cycle ;
\draw  [draw opacity=0][fill={rgb, 255:red, 255; green, 232; blue, 89 }  ,fill opacity=1 ] (73.8,109) -- (133.8,109) -- (133.8,169) -- (73.8,169) -- cycle ;
\draw  [draw opacity=0][dash pattern={on 4.5pt off 4.5pt}] (74.1,109) -- (274,109) -- (274,309) -- (74.1,309) -- cycle ; \draw  [color={rgb, 255:red, 155; green, 155; blue, 155 }  ,draw opacity=1 ][dash pattern={on 4.5pt off 4.5pt}] (74.1,109) -- (74.1,309)(94,109) -- (94,309)(113.9,109) -- (113.9,309)(133.8,109) -- (133.8,309)(153.7,109) -- (153.7,309)(173.6,109) -- (173.6,309)(193.5,109) -- (193.5,309)(213.4,109) -- (213.4,309)(233.3,109) -- (233.3,309)(253.2,109) -- (253.2,309)(273.1,109) -- (273.1,309) ; \draw  [color={rgb, 255:red, 155; green, 155; blue, 155 }  ,draw opacity=1 ][dash pattern={on 4.5pt off 4.5pt}] (74.1,109) -- (274,109)(74.1,128.9) -- (274,128.9)(74.1,148.8) -- (274,148.8)(74.1,168.7) -- (274,168.7)(74.1,188.6) -- (274,188.6)(74.1,208.5) -- (274,208.5)(74.1,228.4) -- (274,228.4)(74.1,248.3) -- (274,248.3)(74.1,268.2) -- (274,268.2)(74.1,288.1) -- (274,288.1)(74.1,308) -- (274,308) ; \draw  [color={rgb, 255:red, 155; green, 155; blue, 155 }  ,draw opacity=1 ][dash pattern={on 4.5pt off 4.5pt}]  ;
\draw  [dash pattern={on 4.5pt off 4.5pt}]  (96,318.99) -- (252,318.54) ;
\draw [shift={(254,318.54)}, rotate = 179.83] [color={rgb, 255:red, 0; green, 0; blue, 0 }  ][line width=0.75]    (10.93,-4.9) .. controls (6.95,-2.3) and (3.31,-0.67) .. (0,0) .. controls (3.31,0.67) and (6.95,2.3) .. (10.93,4.9)   ;
\draw [shift={(94,319)}, rotate = 359.83] [color={rgb, 255:red, 0; green, 0; blue, 0 }  ][line width=0.75]    (10.93,-4.9) .. controls (6.95,-2.3) and (3.31,-0.67) .. (0,0) .. controls (3.31,0.67) and (6.95,2.3) .. (10.93,4.9)   ;
\draw  [dash pattern={on 4.5pt off 4.5pt}]  (64,131.46) -- (64,287) ;
\draw [shift={(64,289)}, rotate = 270] [color={rgb, 255:red, 0; green, 0; blue, 0 }  ][line width=0.75]    (10.93,-4.9) .. controls (6.95,-2.3) and (3.31,-0.67) .. (0,0) .. controls (3.31,0.67) and (6.95,2.3) .. (10.93,4.9)   ;
\draw [shift={(64,129.46)}, rotate = 90] [color={rgb, 255:red, 0; green, 0; blue, 0 }  ][line width=0.75]    (10.93,-4.9) .. controls (6.95,-2.3) and (3.31,-0.67) .. (0,0) .. controls (3.31,0.67) and (6.95,2.3) .. (10.93,4.9)   ;
\draw  [dash pattern={on 0.84pt off 2.51pt}]  (74,319) -- (94,319) ;
\draw [shift={(94,319)}, rotate = 180] [color={rgb, 255:red, 0; green, 0; blue, 0 }  ][line width=0.75]    (0,4.47) -- (0,-4.47)   ;
\draw [shift={(74,319)}, rotate = 180] [color={rgb, 255:red, 0; green, 0; blue, 0 }  ][line width=0.75]    (0,4.47) -- (0,-4.47)   ;
\draw  [dash pattern={on 0.84pt off 2.51pt}]  (64,109) -- (64,129.46) ;
\draw [shift={(64,129.46)}, rotate = 270] [color={rgb, 255:red, 0; green, 0; blue, 0 }  ][line width=0.75]    (0,4.47) -- (0,-4.47)   ;
\draw [shift={(64,109)}, rotate = 270] [color={rgb, 255:red, 0; green, 0; blue, 0 }  ][line width=0.75]    (0,4.47) -- (0,-4.47)   ;
\draw   (710,196.92) -- (747.35,196.92) -- (747.35,185.9) -- (770,207.95) -- (747.35,230) -- (747.35,218.97) -- (710,218.97) -- cycle ;
\draw  [draw opacity=0] (840,160) -- (940,160) -- (940,260) -- (840,260) -- cycle ; \draw   (840,160) -- (840,260)(859.9,160) -- (859.9,260)(879.8,160) -- (879.8,260)(899.7,160) -- (899.7,260)(919.6,160) -- (919.6,260)(939.5,160) -- (939.5,260) ; \draw   (840,160) -- (940,160)(840,179.9) -- (940,179.9)(840,199.8) -- (940,199.8)(840,219.7) -- (940,219.7)(840,239.6) -- (940,239.6)(840,259.5) -- (940,259.5) ; \draw    ;
\draw  [fill={rgb, 255:red, 255; green, 232; blue, 89 }  ,fill opacity=1 ] (840,160) -- (860,160) -- (860,180) -- (840,180) -- cycle ;
\draw  [dash pattern={on 4.5pt off 4.5pt}]  (830,162) -- (830,258) ;
\draw [shift={(830,260)}, rotate = 270] [color={rgb, 255:red, 0; green, 0; blue, 0 }  ][line width=0.75]    (10.93,-4.9) .. controls (6.95,-2.3) and (3.31,-0.67) .. (0,0) .. controls (3.31,0.67) and (6.95,2.3) .. (10.93,4.9)   ;
\draw [shift={(830,160)}, rotate = 90] [color={rgb, 255:red, 0; green, 0; blue, 0 }  ][line width=0.75]    (10.93,-4.9) .. controls (6.95,-2.3) and (3.31,-0.67) .. (0,0) .. controls (3.31,0.67) and (6.95,2.3) .. (10.93,4.9)   ;
\draw  [dash pattern={on 4.5pt off 4.5pt}]  (842,270) -- (938,270) ;
\draw [shift={(940,270)}, rotate = 180] [color={rgb, 255:red, 0; green, 0; blue, 0 }  ][line width=0.75]    (10.93,-4.9) .. controls (6.95,-2.3) and (3.31,-0.67) .. (0,0) .. controls (3.31,0.67) and (6.95,2.3) .. (10.93,4.9)   ;
\draw [shift={(840,270)}, rotate = 0] [color={rgb, 255:red, 0; green, 0; blue, 0 }  ][line width=0.75]    (10.93,-4.9) .. controls (6.95,-2.3) and (3.31,-0.67) .. (0,0) .. controls (3.31,0.67) and (6.95,2.3) .. (10.93,4.9)   ;
\draw  [draw opacity=0][pattern=_oy72wmuew,pattern size=6pt,pattern thickness=0.75pt,pattern radius=0pt, pattern color={rgb, 255:red, 65; green, 117; blue, 5}][dash pattern={on 6.75pt off 4.5pt}][line width=2.25]  (839.8,160) -- (859.9,160) -- (859.9,180.1) -- (839.8,180.1) -- cycle ;
\draw  [dash pattern={on 4.5pt off 4.5pt}]  (81.71,88.97) -- (128.29,61.03) ;
\draw [shift={(130,60)}, rotate = 149.04] [color={rgb, 255:red, 0; green, 0; blue, 0 }  ][line width=0.75]    (10.93,-4.9) .. controls (6.95,-2.3) and (3.31,-0.67) .. (0,0) .. controls (3.31,0.67) and (6.95,2.3) .. (10.93,4.9)   ;
\draw [shift={(80,90)}, rotate = 329.04] [color={rgb, 255:red, 0; green, 0; blue, 0 }  ][line width=0.75]    (10.93,-4.9) .. controls (6.95,-2.3) and (3.31,-0.67) .. (0,0) .. controls (3.31,0.67) and (6.95,2.3) .. (10.93,4.9)   ;
\draw  [dash pattern={on 4.5pt off 4.5pt}]  (831.41,149.59) -- (858.59,122.41) ;
\draw [shift={(860,121)}, rotate = 135] [color={rgb, 255:red, 0; green, 0; blue, 0 }  ][line width=0.75]    (10.93,-4.9) .. controls (6.95,-2.3) and (3.31,-0.67) .. (0,0) .. controls (3.31,0.67) and (6.95,2.3) .. (10.93,4.9)   ;
\draw [shift={(830,151)}, rotate = 315] [color={rgb, 255:red, 0; green, 0; blue, 0 }  ][line width=0.75]    (10.93,-4.9) .. controls (6.95,-2.3) and (3.31,-0.67) .. (0,0) .. controls (3.31,0.67) and (6.95,2.3) .. (10.93,4.9)   ;
\draw  [draw opacity=0][pattern=_7el980ygu,pattern size=6pt,pattern thickness=0.75pt,pattern radius=0pt, pattern color={rgb, 255:red, 189; green, 16; blue, 224}][dash pattern={on 6.75pt off 4.5pt}][line width=2.25]  (870,130) -- (889.55,130) -- (889.55,149.2) -- (870,149.2) -- cycle ;
\draw  [draw opacity=0] (460,230) -- (520,230) -- (520,290) -- (460,290) -- cycle ; \draw   (460,230) -- (460,290)(479.9,230) -- (479.9,290)(499.8,230) -- (499.8,290)(519.7,230) -- (519.7,290) ; \draw   (460,230) -- (520,230)(460,249.9) -- (520,249.9)(460,269.8) -- (520,269.8)(460,289.7) -- (520,289.7) ; \draw    ;
\draw  [draw opacity=0][pattern=_g2c110dgk,pattern size=6pt,pattern thickness=0.75pt,pattern radius=0pt, pattern color={rgb, 255:red, 65; green, 117; blue, 5}][dash pattern={on 6.75pt off 4.5pt}][line width=2.25]  (460,230) -- (520,230) -- (520,290) -- (460,290) -- cycle ;
\draw  [dash pattern={on 4.5pt off 4.5pt}]  (551.41,288.59) -- (628.59,211.41) ;
\draw [shift={(630,210)}, rotate = 135] [color={rgb, 255:red, 0; green, 0; blue, 0 }  ][line width=0.75]    (10.93,-4.9) .. controls (6.95,-2.3) and (3.31,-0.67) .. (0,0) .. controls (3.31,0.67) and (6.95,2.3) .. (10.93,4.9)   ;
\draw [shift={(550,290)}, rotate = 315] [color={rgb, 255:red, 0; green, 0; blue, 0 }  ][line width=0.75]    (10.93,-4.9) .. controls (6.95,-2.3) and (3.31,-0.67) .. (0,0) .. controls (3.31,0.67) and (6.95,2.3) .. (10.93,4.9)   ;
\draw  [dash pattern={on 4.5pt off 4.5pt}]  (451.41,218.59) -- (478.59,191.41) ;
\draw [shift={(480,190)}, rotate = 135] [color={rgb, 255:red, 0; green, 0; blue, 0 }  ][line width=0.75]    (10.93,-4.9) .. controls (6.95,-2.3) and (3.31,-0.67) .. (0,0) .. controls (3.31,0.67) and (6.95,2.3) .. (10.93,4.9)   ;
\draw [shift={(450,220)}, rotate = 315] [color={rgb, 255:red, 0; green, 0; blue, 0 }  ][line width=0.75]    (10.93,-4.9) .. controls (6.95,-2.3) and (3.31,-0.67) .. (0,0) .. controls (3.31,0.67) and (6.95,2.3) .. (10.93,4.9)   ;
\draw  [dash pattern={on 4.5pt off 4.5pt}]  (531.41,139.59) -- (558.59,112.41) ;
\draw [shift={(560,111)}, rotate = 135] [color={rgb, 255:red, 0; green, 0; blue, 0 }  ][line width=0.75]    (10.93,-4.9) .. controls (6.95,-2.3) and (3.31,-0.67) .. (0,0) .. controls (3.31,0.67) and (6.95,2.3) .. (10.93,4.9)   ;
\draw [shift={(530,141)}, rotate = 315] [color={rgb, 255:red, 0; green, 0; blue, 0 }  ][line width=0.75]    (10.93,-4.9) .. controls (6.95,-2.3) and (3.31,-0.67) .. (0,0) .. controls (3.31,0.67) and (6.95,2.3) .. (10.93,4.9)   ;
\draw  [dash pattern={on 4.5pt off 4.5pt}]  (572,110) -- (628,110) ;
\draw [shift={(630,110)}, rotate = 180] [color={rgb, 255:red, 0; green, 0; blue, 0 }  ][line width=0.75]    (10.93,-4.9) .. controls (6.95,-2.3) and (3.31,-0.67) .. (0,0) .. controls (3.31,0.67) and (6.95,2.3) .. (10.93,4.9)   ;
\draw [shift={(570,110)}, rotate = 0] [color={rgb, 255:red, 0; green, 0; blue, 0 }  ][line width=0.75]    (10.93,-4.9) .. controls (6.95,-2.3) and (3.31,-0.67) .. (0,0) .. controls (3.31,0.67) and (6.95,2.3) .. (10.93,4.9)   ;
\draw  [dash pattern={on 4.5pt off 4.5pt}]  (640,122) -- (640,178) ;
\draw [shift={(640,180)}, rotate = 270] [color={rgb, 255:red, 0; green, 0; blue, 0 }  ][line width=0.75]    (10.93,-4.9) .. controls (6.95,-2.3) and (3.31,-0.67) .. (0,0) .. controls (3.31,0.67) and (6.95,2.3) .. (10.93,4.9)   ;
\draw [shift={(640,120)}, rotate = 90] [color={rgb, 255:red, 0; green, 0; blue, 0 }  ][line width=0.75]    (10.93,-4.9) .. controls (6.95,-2.3) and (3.31,-0.67) .. (0,0) .. controls (3.31,0.67) and (6.95,2.3) .. (10.93,4.9)   ;
\draw  [draw opacity=0] (94,128.6) -- (254,128.6) -- (254,288.6) -- (94,288.6) -- cycle ; \draw   (94,128.6) -- (94,288.6)(113.9,128.6) -- (113.9,288.6)(133.8,128.6) -- (133.8,288.6)(153.7,128.6) -- (153.7,288.6)(173.6,128.6) -- (173.6,288.6)(193.5,128.6) -- (193.5,288.6)(213.4,128.6) -- (213.4,288.6)(233.3,128.6) -- (233.3,288.6)(253.2,128.6) -- (253.2,288.6) ; \draw   (94,128.6) -- (254,128.6)(94,148.5) -- (254,148.5)(94,168.4) -- (254,168.4)(94,188.3) -- (254,188.3)(94,208.2) -- (254,208.2)(94,228.1) -- (254,228.1)(94,248) -- (254,248)(94,267.9) -- (254,267.9)(94,287.8) -- (254,287.8) ; \draw    ;

\draw (161,322) node [anchor=north west][inner sep=0.75pt]  [font=\Large] [align=left] {IW};
\draw (31,192) node [anchor=north west][inner sep=0.75pt]  [font=\Large] [align=left] {IH};
\draw (68,327) node [anchor=north west][inner sep=0.75pt]  [font=\large] [align=left] {PW};
\draw (31,107) node [anchor=north west][inner sep=0.75pt]  [font=\large] [align=left] {PH};
\draw (791,192) node [anchor=north west][inner sep=0.75pt]  [font=\Large] [align=left] {OH};
\draw (870,272) node [anchor=north west][inner sep=0.75pt]  [font=\Large] [align=left] {OW};
\draw (93,55) node  [font=\Large] [align=left] {IC};
\draw (822,125) node  [font=\Large] [align=left] {OC};
\draw (394.5,209) node  [font=\Huge] [align=left] {$\displaystyle \circledast $};
\draw (170.5,365) node  [font=\Huge] [align=left] {\textbf{inputs tensor}};
\draw (548,366) node  [font=\Huge] [align=left] {\textbf{weights tensor}};
\draw (892.5,366) node  [font=\Huge] [align=left] {\textbf{outputs tensor}};
\draw (607,259.5) node  [font=\Large] [align=left] {OC};
\draw (450.5,189.5) node  [font=\Large] [align=left] {IC};
\draw (534,114) node  [font=\Large] [align=left] {IC};
\draw (670.5,144) node  [font=\Large] [align=left] {KW};
\draw (598,94) node  [font=\Large] [align=left] {KH};

\end{tikzpicture}

%% file: sketch/im2row.tex
 
\tikzset{
pattern size/.store in=\mcSize, 
pattern size = 5pt,
pattern thickness/.store in=\mcThickness, 
pattern thickness = 0.3pt,
pattern radius/.store in=\mcRadius, 
pattern radius = 1pt}
\makeatletter
\pgfutil@ifundefined{pgf@pattern@name@_4tq49v5vd}{
\pgfdeclarepatternformonly[\mcThickness,\mcSize]{_4tq49v5vd}
{\pgfqpoint{0pt}{0pt}}
{\pgfpoint{\mcSize+\mcThickness}{\mcSize+\mcThickness}}
{\pgfpoint{\mcSize}{\mcSize}}
{
\pgfsetcolor{\tikz@pattern@color}
\pgfsetlinewidth{\mcThickness}
\pgfpathmoveto{\pgfqpoint{0pt}{0pt}}
\pgfpathlineto{\pgfpoint{\mcSize+\mcThickness}{\mcSize+\mcThickness}}
\pgfusepath{stroke}
}}
\makeatother

 
\tikzset{
pattern size/.store in=\mcSize, 
pattern size = 5pt,
pattern thickness/.store in=\mcThickness, 
pattern thickness = 0.3pt,
pattern radius/.store in=\mcRadius, 
pattern radius = 1pt}
\makeatletter
\pgfutil@ifundefined{pgf@pattern@name@_fzb2dr70t}{
\pgfdeclarepatternformonly[\mcThickness,\mcSize]{_fzb2dr70t}
{\pgfqpoint{0pt}{0pt}}
{\pgfpoint{\mcSize+\mcThickness}{\mcSize+\mcThickness}}
{\pgfpoint{\mcSize}{\mcSize}}
{
\pgfsetcolor{\tikz@pattern@color}
\pgfsetlinewidth{\mcThickness}
\pgfpathmoveto{\pgfqpoint{0pt}{0pt}}
\pgfpathlineto{\pgfpoint{\mcSize+\mcThickness}{\mcSize+\mcThickness}}
\pgfusepath{stroke}
}}
\makeatother

 
\tikzset{
pattern size/.store in=\mcSize, 
pattern size = 5pt,
pattern thickness/.store in=\mcThickness, 
pattern thickness = 0.3pt,
pattern radius/.store in=\mcRadius, 
pattern radius = 1pt}
\makeatletter
\pgfutil@ifundefined{pgf@pattern@name@_raj7zqnow}{
\pgfdeclarepatternformonly[\mcThickness,\mcSize]{_raj7zqnow}
{\pgfqpoint{0pt}{0pt}}
{\pgfpoint{\mcSize+\mcThickness}{\mcSize+\mcThickness}}
{\pgfpoint{\mcSize}{\mcSize}}
{
\pgfsetcolor{\tikz@pattern@color}
\pgfsetlinewidth{\mcThickness}
\pgfpathmoveto{\pgfqpoint{0pt}{0pt}}
\pgfpathlineto{\pgfpoint{\mcSize+\mcThickness}{\mcSize+\mcThickness}}
\pgfusepath{stroke}
}}
\makeatother

 
\tikzset{
pattern size/.store in=\mcSize, 
pattern size = 5pt,
pattern thickness/.store in=\mcThickness, 
pattern thickness = 0.3pt,
pattern radius/.store in=\mcRadius, 
pattern radius = 1pt}
\makeatletter
\pgfutil@ifundefined{pgf@pattern@name@_lv680uwdm}{
\pgfdeclarepatternformonly[\mcThickness,\mcSize]{_lv680uwdm}
{\pgfqpoint{0pt}{-\mcThickness}}
{\pgfpoint{\mcSize}{\mcSize}}
{\pgfpoint{\mcSize}{\mcSize}}
{
\pgfsetcolor{\tikz@pattern@color}
\pgfsetlinewidth{\mcThickness}
\pgfpathmoveto{\pgfqpoint{0pt}{\mcSize}}
\pgfpathlineto{\pgfpoint{\mcSize+\mcThickness}{-\mcThickness}}
\pgfusepath{stroke}
}}
\makeatother

 
\tikzset{
pattern size/.store in=\mcSize, 
pattern size = 5pt,
pattern thickness/.store in=\mcThickness, 
pattern thickness = 0.3pt,
pattern radius/.store in=\mcRadius, 
pattern radius = 1pt}
\makeatletter
\pgfutil@ifundefined{pgf@pattern@name@_ue54bsisv}{
\pgfdeclarepatternformonly[\mcThickness,\mcSize]{_ue54bsisv}
{\pgfqpoint{0pt}{-\mcThickness}}
{\pgfpoint{\mcSize}{\mcSize}}
{\pgfpoint{\mcSize}{\mcSize}}
{
\pgfsetcolor{\tikz@pattern@color}
\pgfsetlinewidth{\mcThickness}
\pgfpathmoveto{\pgfqpoint{0pt}{\mcSize}}
\pgfpathlineto{\pgfpoint{\mcSize+\mcThickness}{-\mcThickness}}
\pgfusepath{stroke}
}}
\makeatother

 
\tikzset{
pattern size/.store in=\mcSize, 
pattern size = 5pt,
pattern thickness/.store in=\mcThickness, 
pattern thickness = 0.3pt,
pattern radius/.store in=\mcRadius, 
pattern radius = 1pt}
\makeatletter
\pgfutil@ifundefined{pgf@pattern@name@_gdfqwzare}{
\pgfdeclarepatternformonly[\mcThickness,\mcSize]{_gdfqwzare}
{\pgfqpoint{0pt}{-\mcThickness}}
{\pgfpoint{\mcSize}{\mcSize}}
{\pgfpoint{\mcSize}{\mcSize}}
{
\pgfsetcolor{\tikz@pattern@color}
\pgfsetlinewidth{\mcThickness}
\pgfpathmoveto{\pgfqpoint{0pt}{\mcSize}}
\pgfpathlineto{\pgfpoint{\mcSize+\mcThickness}{-\mcThickness}}
\pgfusepath{stroke}
}}
\makeatother

 
\tikzset{
pattern size/.store in=\mcSize, 
pattern size = 5pt,
pattern thickness/.store in=\mcThickness, 
pattern thickness = 0.3pt,
pattern radius/.store in=\mcRadius, 
pattern radius = 1pt}
\makeatletter
\pgfutil@ifundefined{pgf@pattern@name@_k9seysgqa}{
\pgfdeclarepatternformonly[\mcThickness,\mcSize]{_k9seysgqa}
{\pgfqpoint{0pt}{0pt}}
{\pgfpoint{\mcSize+\mcThickness}{\mcSize+\mcThickness}}
{\pgfpoint{\mcSize}{\mcSize}}
{
\pgfsetcolor{\tikz@pattern@color}
\pgfsetlinewidth{\mcThickness}
\pgfpathmoveto{\pgfqpoint{0pt}{0pt}}
\pgfpathlineto{\pgfpoint{\mcSize+\mcThickness}{\mcSize+\mcThickness}}
\pgfusepath{stroke}
}}
\makeatother

 
\tikzset{
pattern size/.store in=\mcSize, 
pattern size = 5pt,
pattern thickness/.store in=\mcThickness, 
pattern thickness = 0.3pt,
pattern radius/.store in=\mcRadius, 
pattern radius = 1pt}
\makeatletter
\pgfutil@ifundefined{pgf@pattern@name@_iefuit5xn}{
\pgfdeclarepatternformonly[\mcThickness,\mcSize]{_iefuit5xn}
{\pgfqpoint{0pt}{-\mcThickness}}
{\pgfpoint{\mcSize}{\mcSize}}
{\pgfpoint{\mcSize}{\mcSize}}
{
\pgfsetcolor{\tikz@pattern@color}
\pgfsetlinewidth{\mcThickness}
\pgfpathmoveto{\pgfqpoint{0pt}{\mcSize}}
\pgfpathlineto{\pgfpoint{\mcSize+\mcThickness}{-\mcThickness}}
\pgfusepath{stroke}
}}
\makeatother
\tikzset{every picture/.style={line width=0.75pt}} 

\begin{tikzpicture}[x=0.75pt,y=0.75pt,yscale=-1,xscale=1]

\draw  [draw opacity=0] (900,79.25) -- (1000,79.25) -- (1000,179.25) -- (900,179.25) -- cycle ; \draw   (900,79.25) -- (900,179.25)(919.9,79.25) -- (919.9,179.25)(939.8,79.25) -- (939.8,179.25)(959.7,79.25) -- (959.7,179.25)(979.6,79.25) -- (979.6,179.25)(999.5,79.25) -- (999.5,179.25) ; \draw   (900,79.25) -- (1000,79.25)(900,99.15) -- (1000,99.15)(900,119.05) -- (1000,119.05)(900,138.95) -- (1000,138.95)(900,158.85) -- (1000,158.85)(900,178.75) -- (1000,178.75) ; \draw    ;
\draw  [draw opacity=0][fill={rgb, 255:red, 255; green, 255; blue, 255 }  ,fill opacity=1 ] (880.15,109) -- (969.55,109) -- (969.55,200.2) -- (880.15,200.2) -- cycle ;
\draw  [fill={rgb, 255:red, 255; green, 232; blue, 89 }  ,fill opacity=1 ] (381,61.2) -- (560.1,61.2) -- (560.1,81.1) -- (381,81.1) -- cycle ;
\draw  [fill={rgb, 255:red, 255; green, 232; blue, 89 }  ,fill opacity=1 ] (110.1,99.3) -- (169.8,99.3) -- (169.8,159.2) -- (110.1,159.2) -- cycle ;
\draw  [draw opacity=0][dash pattern={on 4.5pt off 4.5pt}] (110.1,99.5) -- (250,99.5) -- (250,239) -- (110.1,239) -- cycle ; \draw  [color={rgb, 255:red, 155; green, 155; blue, 155 }  ,draw opacity=1 ][dash pattern={on 4.5pt off 4.5pt}] (110.1,99.5) -- (110.1,239)(130,99.5) -- (130,239)(149.9,99.5) -- (149.9,239)(169.8,99.5) -- (169.8,239)(189.7,99.5) -- (189.7,239)(209.6,99.5) -- (209.6,239)(229.5,99.5) -- (229.5,239)(249.4,99.5) -- (249.4,239) ; \draw  [color={rgb, 255:red, 155; green, 155; blue, 155 }  ,draw opacity=1 ][dash pattern={on 4.5pt off 4.5pt}] (110.1,99.5) -- (250,99.5)(110.1,119.4) -- (250,119.4)(110.1,139.3) -- (250,139.3)(110.1,159.2) -- (250,159.2)(110.1,179.1) -- (250,179.1)(110.1,199) -- (250,199)(110.1,218.9) -- (250,218.9)(110.1,238.8) -- (250,238.8) ; \draw  [color={rgb, 255:red, 155; green, 155; blue, 155 }  ,draw opacity=1 ][dash pattern={on 4.5pt off 4.5pt}]  ;
\draw  [draw opacity=0] (130,119.5) -- (230,119.5) -- (230,219.5) -- (130,219.5) -- cycle ; \draw   (130,119.5) -- (130,219.5)(149.9,119.5) -- (149.9,219.5)(169.8,119.5) -- (169.8,219.5)(189.7,119.5) -- (189.7,219.5)(209.6,119.5) -- (209.6,219.5)(229.5,119.5) -- (229.5,219.5) ; \draw   (130,119.5) -- (230,119.5)(130,139.4) -- (230,139.4)(130,159.3) -- (230,159.3)(130,179.2) -- (230,179.2)(130,199.1) -- (230,199.1)(130,219) -- (230,219) ; \draw    ;
\draw  [dash pattern={on 4.5pt off 4.5pt}]  (132,249) -- (228,249) ;
\draw [shift={(230,249)}, rotate = 180] [color={rgb, 255:red, 0; green, 0; blue, 0 }  ][line width=0.75]    (10.93,-4.9) .. controls (6.95,-2.3) and (3.31,-0.67) .. (0,0) .. controls (3.31,0.67) and (6.95,2.3) .. (10.93,4.9)   ;
\draw [shift={(130,249)}, rotate = 0] [color={rgb, 255:red, 0; green, 0; blue, 0 }  ][line width=0.75]    (10.93,-4.9) .. controls (6.95,-2.3) and (3.31,-0.67) .. (0,0) .. controls (3.31,0.67) and (6.95,2.3) .. (10.93,4.9)   ;
\draw  [dash pattern={on 4.5pt off 4.5pt}]  (90,122) -- (90,217.04) ;
\draw [shift={(90,219.04)}, rotate = 270] [color={rgb, 255:red, 0; green, 0; blue, 0 }  ][line width=0.75]    (10.93,-4.9) .. controls (6.95,-2.3) and (3.31,-0.67) .. (0,0) .. controls (3.31,0.67) and (6.95,2.3) .. (10.93,4.9)   ;
\draw [shift={(90,120)}, rotate = 90] [color={rgb, 255:red, 0; green, 0; blue, 0 }  ][line width=0.75]    (10.93,-4.9) .. controls (6.95,-2.3) and (3.31,-0.67) .. (0,0) .. controls (3.31,0.67) and (6.95,2.3) .. (10.93,4.9)   ;
\draw   (270,158.53) -- (311.23,158.53) -- (311.23,150) -- (330,167.05) -- (311.23,184.1) -- (311.23,175.58) -- (270,175.58) -- cycle ;
\draw  [draw opacity=0] (870.05,109) -- (970.05,109) -- (970.05,209) -- (870.05,209) -- cycle ; \draw   (870.05,109) -- (870.05,209)(889.95,109) -- (889.95,209)(909.85,109) -- (909.85,209)(929.75,109) -- (929.75,209)(949.65,109) -- (949.65,209)(969.55,109) -- (969.55,209) ; \draw   (870.05,109) -- (970.05,109)(870.05,128.9) -- (970.05,128.9)(870.05,148.8) -- (970.05,148.8)(870.05,168.7) -- (970.05,168.7)(870.05,188.6) -- (970.05,188.6)(870.05,208.5) -- (970.05,208.5) ; \draw    ;
\draw  [fill={rgb, 255:red, 255; green, 232; blue, 89 }  ,fill opacity=1 ] (870.05,109) -- (890.05,109) -- (890.05,129) -- (870.05,129) -- cycle ;
\draw  [dash pattern={on 4.5pt off 4.5pt}]  (860.05,111) -- (860.05,207) ;
\draw [shift={(860.05,209)}, rotate = 270] [color={rgb, 255:red, 0; green, 0; blue, 0 }  ][line width=0.75]    (10.93,-4.9) .. controls (6.95,-2.3) and (3.31,-0.67) .. (0,0) .. controls (3.31,0.67) and (6.95,2.3) .. (10.93,4.9)   ;
\draw [shift={(860.05,109)}, rotate = 90] [color={rgb, 255:red, 0; green, 0; blue, 0 }  ][line width=0.75]    (10.93,-4.9) .. controls (6.95,-2.3) and (3.31,-0.67) .. (0,0) .. controls (3.31,0.67) and (6.95,2.3) .. (10.93,4.9)   ;
\draw  [dash pattern={on 4.5pt off 4.5pt}]  (872.05,219) -- (968.05,219) ;
\draw [shift={(970.05,219)}, rotate = 180] [color={rgb, 255:red, 0; green, 0; blue, 0 }  ][line width=0.75]    (10.93,-4.9) .. controls (6.95,-2.3) and (3.31,-0.67) .. (0,0) .. controls (3.31,0.67) and (6.95,2.3) .. (10.93,4.9)   ;
\draw [shift={(870.05,219)}, rotate = 0] [color={rgb, 255:red, 0; green, 0; blue, 0 }  ][line width=0.75]    (10.93,-4.9) .. controls (6.95,-2.3) and (3.31,-0.67) .. (0,0) .. controls (3.31,0.67) and (6.95,2.3) .. (10.93,4.9)   ;
\draw  [draw opacity=0][pattern=_4tq49v5vd,pattern size=6pt,pattern thickness=0.75pt,pattern radius=0pt, pattern color={rgb, 255:red, 65; green, 117; blue, 5}][dash pattern={on 6.75pt off 4.5pt}][line width=2.25]  (890.05,109) -- (910.15,109) -- (910.15,129.1) -- (890.05,129.1) -- cycle ;
\draw  [draw opacity=0][pattern=_fzb2dr70t,pattern size=6pt,pattern thickness=0.75pt,pattern radius=0pt, pattern color={rgb, 255:red, 65; green, 117; blue, 5}][dash pattern={on 6.75pt off 4.5pt}][line width=2.25]  (130,99.6) -- (189.7,99.6) -- (189.7,159.3) -- (130,159.3) -- cycle ;
\draw  [draw opacity=0] (381,61) -- (561,61) -- (561,251) -- (381,251) -- cycle ; \draw   (381,61) -- (381,251)(400.9,61) -- (400.9,251)(420.8,61) -- (420.8,251)(440.7,61) -- (440.7,251)(460.6,61) -- (460.6,251)(480.5,61) -- (480.5,251)(500.4,61) -- (500.4,251)(520.3,61) -- (520.3,251)(540.2,61) -- (540.2,251)(560.1,61) -- (560.1,251) ; \draw   (381,61) -- (561,61)(381,80.9) -- (561,80.9)(381,100.8) -- (561,100.8)(381,120.7) -- (561,120.7)(381,140.6) -- (561,140.6)(381,160.5) -- (561,160.5)(381,180.4) -- (561,180.4)(381,200.3) -- (561,200.3)(381,220.2) -- (561,220.2)(381,240.1) -- (561,240.1) ; \draw    ;
\draw   (760.05,142.5) -- (787.5,142.5) -- (787.5,130) -- (810,155) -- (787.5,180) -- (787.5,167.5) -- (760.05,167.5) -- cycle ;
\draw  [draw opacity=0][pattern=_raj7zqnow,pattern size=6pt,pattern thickness=0.75pt,pattern radius=0pt, pattern color={rgb, 255:red, 65; green, 117; blue, 5}][dash pattern={on 6.75pt off 4.5pt}][line width=2.25]  (380.8,80.9) -- (560.1,80.9) -- (560.1,101.2) -- (380.8,101.2) -- cycle ;
\draw   (634.77,145.77) -- (640.48,151.48) -- (630.96,161) -- (640.48,170.52) -- (634.52,176.48) -- (625,166.96) -- (615.48,176.48) -- (609.77,170.77) -- (619.3,161.25) -- (609.77,151.73) -- (615.73,145.77) -- (625.25,155.3) -- cycle ;
\draw  [draw opacity=0] (701,70) -- (740.8,70) -- (740.8,249.1) -- (701,249.1) -- cycle ; \draw   (720.9,70) -- (720.9,249.1) ; \draw   (701,89.9) -- (740.8,89.9)(701,109.8) -- (740.8,109.8)(701,129.7) -- (740.8,129.7)(701,149.6) -- (740.8,149.6)(701,169.5) -- (740.8,169.5)(701,189.4) -- (740.8,189.4)(701,209.3) -- (740.8,209.3)(701,229.2) -- (740.8,229.2) ; \draw   (701,70) -- (740.8,70) -- (740.8,249.1) -- (701,249.1) -- cycle ;
\draw  [draw opacity=0][pattern=_lv680uwdm,pattern size=6pt,pattern thickness=0.75pt,pattern radius=0pt, pattern color={rgb, 255:red, 189; green, 16; blue, 224}][dash pattern={on 6.75pt off 4.5pt}][line width=2.25]  (149.9,99.6) -- (209.6,99.6) -- (209.6,159.3) -- (149.9,159.3) -- cycle ;
\draw  [draw opacity=0][pattern=_ue54bsisv,pattern size=6pt,pattern thickness=0.75pt,pattern radius=0pt, pattern color={rgb, 255:red, 189; green, 16; blue, 224}][dash pattern={on 6.75pt off 4.5pt}][line width=2.25]  (381,100.4) -- (560.3,100.4) -- (560.3,120.7) -- (381,120.7) -- cycle ;
\draw  [draw opacity=0][pattern=_gdfqwzare,pattern size=6pt,pattern thickness=0.75pt,pattern radius=0pt, pattern color={rgb, 255:red, 189; green, 16; blue, 224}][dash pattern={on 6.75pt off 4.5pt}][line width=2.25]  (909.85,108.6) -- (929.55,108.6) -- (929.55,128.9) -- (909.85,128.9) -- cycle ;
\draw  [dash pattern={on 4.5pt off 4.5pt}]  (691,72) -- (691,248) ;
\draw [shift={(691,250)}, rotate = 270] [color={rgb, 255:red, 0; green, 0; blue, 0 }  ][line width=0.75]    (10.93,-4.9) .. controls (6.95,-2.3) and (3.31,-0.67) .. (0,0) .. controls (3.31,0.67) and (6.95,2.3) .. (10.93,4.9)   ;
\draw [shift={(691,70)}, rotate = 90] [color={rgb, 255:red, 0; green, 0; blue, 0 }  ][line width=0.75]    (10.93,-4.9) .. controls (6.95,-2.3) and (3.31,-0.67) .. (0,0) .. controls (3.31,0.67) and (6.95,2.3) .. (10.93,4.9)   ;
\draw  [dash pattern={on 4.5pt off 4.5pt}]  (559,51) -- (383,51) ;
\draw [shift={(381,51)}, rotate = 360] [color={rgb, 255:red, 0; green, 0; blue, 0 }  ][line width=0.75]    (10.93,-4.9) .. controls (6.95,-2.3) and (3.31,-0.67) .. (0,0) .. controls (3.31,0.67) and (6.95,2.3) .. (10.93,4.9)   ;
\draw [shift={(561,51)}, rotate = 180] [color={rgb, 255:red, 0; green, 0; blue, 0 }  ][line width=0.75]    (10.93,-4.9) .. controls (6.95,-2.3) and (3.31,-0.67) .. (0,0) .. controls (3.31,0.67) and (6.95,2.3) .. (10.93,4.9)   ;
\draw  [dash pattern={on 4.5pt off 4.5pt}]  (371,258) -- (371,62) ;
\draw [shift={(371,60)}, rotate = 90] [color={rgb, 255:red, 0; green, 0; blue, 0 }  ][line width=0.75]    (10.93,-4.9) .. controls (6.95,-2.3) and (3.31,-0.67) .. (0,0) .. controls (3.31,0.67) and (6.95,2.3) .. (10.93,4.9)   ;
\draw [shift={(371,260)}, rotate = 270] [color={rgb, 255:red, 0; green, 0; blue, 0 }  ][line width=0.75]    (10.93,-4.9) .. controls (6.95,-2.3) and (3.31,-0.67) .. (0,0) .. controls (3.31,0.67) and (6.95,2.3) .. (10.93,4.9)   ;
\draw  [dash pattern={on 4.5pt off 4.5pt}]  (701,60) -- (740,60) ;
\draw [shift={(740,60)}, rotate = 180] [color={rgb, 255:red, 0; green, 0; blue, 0 }  ][line width=0.75]    (0,5.59) -- (0,-5.59)   ;
\draw [shift={(701,60)}, rotate = 180] [color={rgb, 255:red, 0; green, 0; blue, 0 }  ][line width=0.75]    (0,5.59) -- (0,-5.59)   ;
\draw    (571,60) .. controls (592.17,100.5) and (590.17,207.5) .. (571,250) ;
\draw    (50,60) .. controls (33.17,101.5) and (33.17,201.5) .. (50,250) ;
\draw  [dash pattern={on 4.5pt off 4.5pt}]  (898.56,71.39) -- (870.49,98.61) ;
\draw [shift={(869.05,100)}, rotate = 315.89] [color={rgb, 255:red, 0; green, 0; blue, 0 }  ][line width=0.75]    (10.93,-4.9) .. controls (6.95,-2.3) and (3.31,-0.67) .. (0,0) .. controls (3.31,0.67) and (6.95,2.3) .. (10.93,4.9)   ;
\draw [shift={(900,70)}, rotate = 135.89] [color={rgb, 255:red, 0; green, 0; blue, 0 }  ][line width=0.75]    (10.93,-4.9) .. controls (6.95,-2.3) and (3.31,-0.67) .. (0,0) .. controls (3.31,0.67) and (6.95,2.3) .. (10.93,4.9)   ;
\draw  [draw opacity=0][pattern=_k9seysgqa,pattern size=6pt,pattern thickness=0.75pt,pattern radius=0pt, pattern color={rgb, 255:red, 65; green, 117; blue, 5}][dash pattern={on 6.75pt off 4.5pt}][line width=2.25]  (919.7,79.05) -- (939.8,79.05) -- (939.8,99.15) -- (919.7,99.15) -- cycle ;
\draw  [draw opacity=0][pattern=_iefuit5xn,pattern size=6pt,pattern thickness=0.75pt,pattern radius=0pt, pattern color={rgb, 255:red, 189; green, 16; blue, 224}][dash pattern={on 6.75pt off 4.5pt}][line width=2.25]  (940,78.85) -- (959.7,78.85) -- (959.7,99.15) -- (940,99.15) -- cycle ;
\draw  [fill={rgb, 255:red, 255; green, 232; blue, 89 }  ,fill opacity=1 ] (900,79.15) -- (920,79.15) -- (920,99.15) -- (900,99.15) -- cycle ;
\draw  [dash pattern={on 4.5pt off 4.5pt}]  (91.79,99.11) -- (128.21,80.89) ;
\draw [shift={(130,80)}, rotate = 153.43] [color={rgb, 255:red, 0; green, 0; blue, 0 }  ][line width=0.75]    (10.93,-4.9) .. controls (6.95,-2.3) and (3.31,-0.67) .. (0,0) .. controls (3.31,0.67) and (6.95,2.3) .. (10.93,4.9)   ;
\draw [shift={(90,100)}, rotate = 333.43] [color={rgb, 255:red, 0; green, 0; blue, 0 }  ][line width=0.75]    (10.93,-4.9) .. controls (6.95,-2.3) and (3.31,-0.67) .. (0,0) .. controls (3.31,0.67) and (6.95,2.3) .. (10.93,4.9)   ;

\draw (175,265) node  [font=\Large] [align=left] {IW};
\draw (67,165) node  [font=\Large] [align=left] {IH};
\draw (821,142) node [anchor=north west][inner sep=0.75pt]  [font=\Large] [align=left] {OH};
\draw (905.05,231) node [anchor=north west][inner sep=0.75pt]   [align=left] {OW};
\draw (300,135) node  [font=\Large] [align=left] {\textbf{im2row}};
\draw (176.45,295) node  [font=\huge] [align=left] {\textbf{src tensor}};
\draw (478.45,294) node  [font=\huge] [align=left] {\textbf{im2row buffer}};
\draw (713.45,294) node  [font=\huge] [align=left] {\textbf{weights tensor}};
\draw (918.5,294) node  [font=\huge] [align=left] {\textbf{output tensor}};
\draw (625,130.5) node  [font=\Large] [align=left] {\textbf{GEMM}};
\draw (675,158) node  [font=\Large,rotate=-270] [align=left] {KH x KW x IC};
\draw (472,35) node  [font=\Large] [align=left] {KH x KW x IC};
\draw (355,155) node  [font=\Large,rotate=-270] [align=left] {MB x IH x IW};
\draw (722,45) node  [font=\Large] [align=left] {OC};
\draw (851,62) node [anchor=north west][inner sep=0.75pt]  [font=\Large] [align=left] {OC};
\draw (85,75) node  [font=\Large] [align=left] {IC=1};

\end{tikzpicture}